\documentclass[12pt]{iopart}

\usepackage{iopams}

\usepackage{cite}
\usepackage{graphicx}
\usepackage{url}

\usepackage{bm}

\usepackage{color}

\begin{document}

%\title{Model Compression of Neural Networks with Koopman Operator}
\title[Extraction of nonlinearity in neural networks with Koopman operator]{Extraction of nonlinearity in neural networks with Koopman operator}

\author{Naoki Sugishita, Kayo Kinjo, and Jun Ohkubo}
\address{Graduate School of Science and Engineering, Saitama University, \\
255 Shimo-Okubo, Sakura-ku, Saitama 338-8570, Japan}
\ead{johkubo@mail.saitama-u.ac.jp}
\vspace{10pt}
\begin{indented}
\item[]
\end{indented}

\begin{abstract}
Nonlinearity plays a crucial role in deep neural networks. In this paper, we investigate the degree to which the nonlinearity of the neural network is essential. For this purpose, we employ the Koopman operator, extended dynamic mode decomposition, and the tensor-train format. The Koopman operator approach has been recently developed in physics and nonlinear sciences; the Koopman operator deals with the time evolution in the observable space instead of the state space. Since we can replace the nonlinearity in the state space with the linearity in the observable space, it is a hopeful candidate for understanding complex behavior in nonlinear systems. Here, we analyze learned neural networks for the classification problems. As a result, the replacement of the nonlinear middle layers with the Koopman matrix yields enough accuracy in numerical experiments. In addition, we confirm that the pruning of the Koopman matrix gives sufficient accuracy even at high compression ratios. These results indicate the possibility of extracting some features in the neural networks with the Koopman operator approach.
\end{abstract}

%\maketitle

\section{\label{sec:intro}Introduction}

As is well known, deep learning performs well in a wide range of research and application areas \cite{LeCun2015,Prince2023,Bishop2023}. Exploitation of nonlinearities can produce highly accurate outputs. However, one of the problems with deep learning is that the number of parameters is enormous. In order to run such networks in environments with limited computational resources, model compression methods have been studied to reduce the number of model parameters. In fact, there are many studies that aim to compress neural networks \cite{Liang_2021_PruningSurvey,Cheng2023}. Quantization and pruning are typical examples of model compression methods. Quantization reduces the data size of parameters by reducing the number of bits required to represent floating-point variables on a computer. Pruning reduces the number of connections between nodes; weights with small contributions are set to zero. These methods often require hardware support to maintain high accuracy while reducing memory consumption at runtime \cite{Gale_2020_GPUKernel}. There are other types of pruning, i.e., the spectral pruning \cite{Giambagli2021,Buffoni2022}. Other model compression methods in recent years include tensor decompositions with a low-rank approximation for the convolutional layer of a neural network \cite{Tai_2016_conv_tensor, Kim_2016_conv_tensor, Aggarwal_2018_conv_tensor}.

Here, we raise the following question from a physics point of view: Are the effects of nonlinearity exploited fully, or are only some parts of nonlinearity enough in performance? In the extreme case, if nonlinearity plays no role, one can use linear algebra and matrix multiplication. Of course, we know that nonlinearity is essential. Is there room left for linear algebra to enter neural networks?

Recently, the Koopman operator has attracted attention in various fields. The Koopman operator \cite{Koopman_1931} is defined in a dynamical system; it deals with the time evolution in the observable space, not the state space. While the observable space is infinite-dimensional, the Koopman operator is linear even if the underlying dynamical system is nonlinear. It is worth noting that several methods have been developed in recent years to obtain Koopman operators from an observed data set. One of the methods is dynamic mode decomposition (DMD), which analyzes dynamical systems from observed data \cite{Rowley_2009_DMDKoopmam}. Extended methods such as extended dynamic mode decomposition (EDMD) can efficiently approximate the Koopman operator as a finite-dimensional matrix \cite{Williams_2015_EDMD}; the development of EDMD made it possible to precisely analyze dynamical systems with only a data set, without any special knowledge of the governing equations. Applications of EDMD are currently being studied, including time series prediction, system identification \cite{Mauroy_2020_KoopmanBasedLT}, and control \cite{Korda_2018_KoopmanMPC}. For more details, see the recent review paper \cite{Brunton2022}.

Some studies also combined neural networks with Koopman operator theory \cite{Li_2017_DicLearn, Takeishi_2017_DicLearn, Lusch_2018_DLforKoopmanEmbed, Menga_2024}; the basic idea is to employ the neural networks to construct dictionaries for the Koopman operator. Actually, some works reported that time evolution predictors with the Koopman operator show better performance in time series prediction tasks than conventional neural networks such as recurrent neural networks \cite{Azencot_2022_ConsistentKoopmanAutoencooder}. Another way to apply Koopman operator theory to neural networks is to consider the learning process of weights as a dynamical system. For example, in \cite{Dogra_2020_OptNNwithKoopmam}, the Koopman operator is constructed from the early-stage learning trajectory of weights. Then, the use of the constructed Koopman operator in the subsequent optimization process significantly reduces the learning time. Reference \cite{Redman_2022_PruningNNwithKoopman} discussed the pruning methods in a unified manner with the aid of the Koopman operator theory. Reference \cite{Konishi_2023_koopman_nn_learn} reported that good performance is achieved by applying the Koopman theory to the learning process of a network called the deep equilibrium model (DEQ).

The interpretability of neural networks is also a hot topic. There are some types of approach, such as a circuit approach\cite{Conmy2023} and a differential equation approach\cite{Marino2023}. The connection of the neural network with the dynamical system stems the ResNet \cite{He2016}, and nowadays, many studies consider the internal process of neural networks as dynamical systems. For example, reference \cite{Lu_2018_NN2ODE} considered the ResNet as a discretization of ordinary differential equations by the Euler's method and shows the improvement of the accuracy of classification problems with the aid of another discretization method. Reference \cite{Chen_2018_NODE} proposed a network called neural ODE, whose internal architecture is the ordinary differential equation; this fact yields straightforwardly to apply existing time evolution solvers in the forward propagation stage. In \cite{Chang_2018_antisymmetricrnn}, the recurrent neural network is regarded as a discretized ordinary differential equation. Hence, it is possible to apply the theory of stability of dynamical systems to the problems of gradient vanishing and explosion during training of the recurrent neural network. These discussions yielded a new structure of the recurrent neural network that avoids gradient vanishing and explosion; the proposed network showed better results than existing methods such as long short-term memory (LSTM) networks. Furthermore, the viewpoint of the dynamical systems, including partial differential equations, has also been studied; for example, see \cite{E2017,Han2018,Ruthotto2020}. One would expect that discussions based on the dynamical systems will provide a better understanding of deep learning in the future.

As introduced above, some studies consider the interior of a neural network as a dynamical system. However, few studies have used the Koopman operator theory for such dynamical systems and replaced the intermediate layer with the corresponding approximated Koopman operator, i.e., the corresponding Koopman matrix. If performance is sufficient with only limited nonlinearity, a simple calculation with linear algebra will contribute to compressing neural networks.

In this study, we discuss the effects of nonlinearity in trained neural networks. Of course, it is difficult to investigate the nonlinear effects, and the naive EDMD can deal with only a limited number of variables because of the curse of dimensionality; the memory size increases exponentially with the number of variables. Hence, we here employ the so-called tensor-train (TT) format \cite{Oseledets2009a,Oseledets2009b,Oseledets2011}, which expands the scope of application of DMD and EDMD (for example, see~\cite{Klus2018,Gelß2019}). After the preliminary investigation for nonlinearity, we consider neural networks for classification problems as discrete-time dynamical systems and perform further analysis. Then, we discuss the extent to which Koopman operators extract the essence of neural networks.

The remaining part of the present paper is as follows. Section~2 describes the existing theory and methods necessary for the proposed method. In section~3, we propose a model replacement framework with the aid of the Koopman operator. Preliminary numerical investigations using the TT format are also given. The main discussion with the proposed framework is given in section~4. Section~5 gives concluding remarks.

\section{Prerequisite}

The Koopman operator is a time evolution operator obtained by considering the time evolution of a dynamical system in a function space, called the observable space. Specifically, we first consider the following discrete-time dynamical system:
\begin{equation}
\bm{x}_{t+1} = F(\bm{x}_t), \quad \bm{x}_t \in \mathbb{R}^D.
\end{equation}
The observable space $\mathcal{F}$ is a set of scalar-valued functions, which are called observables. The observables take the state variable $\bm x_t$ as input. Hence, the space $\mathcal{F}$ is given as
\begin{equation}
\mathcal{F} = \{ f | f: \mathbb{R}^D \rightarrow \mathbb{R} \}.
\end{equation}
The Koopman operator $\mathcal K$ is an operator on this observable space, defined by
\begin{equation}
(\mathcal{K} f)(\bm{x}_t) = f(\bm{x}_{t+1}). \label{eq:koopman_def}
\end{equation}
The action of the Koopman operator in \eref{eq:koopman_def} changes an observable to another observable that outputs the function value after time evolution. For any constants $a, b \in \mathbb R$ and observables $f, g \in \mathcal F$, we have
\begin{eqnarray}
\left\{ \mathcal{K}(af + bg) \right\} (\bm{x}_t) &= (af+bg)(\bm{x}_{t+1}) \nonumber\\
&= af(\bm{x}_{t+1}) + bg(\bm{x}_{t+1}) \nonumber\\
&= a(\mathcal{K}f)(\bm{x}_t) + b(\mathcal{K}g)(\bm{x}_t) \nonumber\\
&= (a\mathcal{K}f + b\mathcal{K}g)(\bm{x}_t).
\end{eqnarray}
Therefore, the Koopman operator is a linear operator.

The observable space is the space of functions, and its dimension is infinite. Hence, we cannot handle the Koopman operator numerically. One method for numerically approximating the Koopman operator is the EDMD algorithm \cite{Williams_2015_EDMD}. The EDMD algorithm computes an approximation of the Koopman operator based on pairs of the time evolution of state variables $\{(\bm{x}_i, \bm{y}_i)\}_{i=1}^M = \{(\bm{x}_i, F(\bm{x}_i))\}_{i=1}^M$, which are called snapshot pairs. The calculation method starts with approximating an infinite dimensional observable space with a space spanned by a finite number of basis functions $\{\phi_j\}_{j=1}^L$, i.e., $\mathcal{F}_{\bm{\Phi}} = \mathrm{span}(\{\phi_i\}_{j=1}^L) \subset \mathcal F$. The basis functions are called dictionary functions and are sometimes redundant.

Since the Koopman operator is a linear operator, let $f=\sum_{i=j}^L a_j \phi_j = \bm{a}^\top \bm{\Phi} \in \mathcal{F}_{\bm{\Phi}}$. Then, the approximated Koopman operator on $\mathcal{F}_{\bm{\Phi}}$ can be expressed by the matrix $K \in \mathbb{R}^{L \times L}$, which is called the Koopman matrix:
\begin{equation}
\mathcal{K} f \approx (K \bm{a})^\top \bm{\Phi}.
\end{equation}
The Koopman matrix is obtained from the snapshot pairs by minimizing the following cost function:
\begin{equation}
\ell(K) = \sum_{i=1}^M \left\| \bm{\Phi}(\bm{y}_i) - K^\top \bm{\Phi}(\bm{x}_i) \right\|^2.
 \label{eq:edmd_argmin}
\end{equation}
This cost function represents the empirical mean squared error between the true values of the time-evolved dictionary functions and the corresponding predicted values on each snapshot pair. In \eref{eq:edmd_argmin}, $\bm{\Phi}(\bm{x}_i)$ represents that a data vector $\bm{x}_i$ is converted to a higher dimensional vector constructed by the dictionary functions. This operation is sometimes called ``lifting''. Minimizing the cost function $\ell(K)$ is a linear least-squares problem whose solution is given by
\begin{equation} \label{eq:koopmanmat}
K = \bm{\Phi}(X)^\dagger \bm{\Phi}(Y),
\end{equation}
where ${}^\dagger$ means the pseudo-inverse matrix. $\bm{\Phi}(X)$ and $\bm{\Phi}(Y) \in \mathbb{R}^{M \times L}$ are the following matrices of the lifted data points:
\begin{equation}
\bm\Phi(X) =
\left(
\bm\Phi(\bm x_1) \ \cdots \ \bm\Phi(\bm x_M)
\right)^\top
, \quad
\bm\Phi(Y) =
\left(
\bm\Phi(\bm y_1) \ \cdots \ \bm\Phi(\bm y_M)
\right)^\top.
\end{equation}

By including the identity map $f_{\mathrm{id}}(\bm{x}_t)=\bm{x}_t$ in the space of dictionary functions $\mathcal{F}_{\bm \Phi}$, the time evolution of the system is approximated by the EDMD algorithm. The identity map $f_{\mathrm{id}}(\bm{x}_t)$ is a vector-valued function, and we define that the Koopman operator $\mathcal{K}$ acts elementwise on $f_{\mathrm{id}}(\bm{x}_t)$. Then, if there exists a coefficient matrix $B \in \mathbb{R}^{L \times D}$ such that $B^\top \bm{\Phi}(\bm{x}_t) = \bm{x}_t$, we have
\begin{equation} \label{eq:evolve_approx}
\bm{x}_{t+1} = (\mathcal{K} f_{\mathrm{id}})(\bm{x}_t) \approx (KB)^\top \bm{\Phi}(\bm{x}_t).
\end{equation}
In the EDMD algorithm, the size of the dictionary should be large enough to approximate the observable functions correctly. Hence, $L \gg D$ in general.

\section{Replacement with Koopman operator and investigations of nonlinearity}

In this section, we propose a replacement of the intermediate layers in neural networks with the Koopman matrix. After explaining the neural network architecture, we perform a preliminary numerical investigation in which the effects of nonlinearity are checked with the TT format.

\subsection{Overview of the proposed method}

% ここに、目的を調べるためにこのような方法を提案する、という形を追加。

As introduced in section~1, many studies consider the internal process of neural networks as a dynamical system. Although the dynamical system is nonlinear, we can apply the EDMD algorithm in principle. If the EDMD algorithm captures the essential part of the dynamical system, we can deal with this part with a linear transformation using a matrix, albeit in an approximate observable space. Then, we consider the intermediate process in the neural networks as a discrete-time dynamical system and replace it with only a matrix.  To investigate the possibility of extracting the essential features with the Koopman operator approach, we propose the following procedure:
\begin{enumerate}
\item Prepare a trained neural network and a training dataset for the model compression process.
\item Generate snapshot pairs by taking two intermediate states of the trained neural network with the prepared training dataset.
\item Calculate the Koopman matrix by the EDMD algorithm using the generated snapshot pairs.
\item Compress the obtained Koopman matrix multiplied by the coefficient matrix $B$ in \eref{eq:evolve_approx} with the aid of the singular value decomposition.
\item ``Mimic'' the original network by replacing a part of the neural network with the compressed Koopman matrix.
\end{enumerate}
Detailed conditions of the target neural network and the method of generating snapshot pairs are explained later in section~3.2.

To compress the matrix in (iv) in the above procedure, we first extract only the part necessary to mimic the original network from the Koopman matrix. Specifically, the matrix $A = KB \in \mathbb{R}^{L \times D}$ is obtained by multiplying the Koopman matrix $K$ by the coefficient matrix $B$ that represents the identity map in the space of dictionary functions. Note that if the identity map itself is included as a dictionary function, we only need to extract the columns of the corresponding Koopman matrix.

Secondly, we perform the singular value decomposition on the matrix $A = KB$ as follows:
\begin{equation}
A = U \Sigma V^\top,
\end{equation}
where $U \in \mathbb{R}^{L \times L}$ and $V \in \mathbb{R}^{D \times D}$ are orthogonal matrices, and $\Sigma \in \mathbb{R}^{L \times D}$ is a block matrix whose upper left part is a diagonal matrix of singular values $\sigma_1, \ldots, \sigma_r > 0 \ ( r=\mathrm{rank}\ A)$ and the rest parts are zero matrices. Then, we select $s$ singular values obtained by the singular value decomposition in descending order and extract the elements related to them from the matrices after singular value decomposition. If a sufficiently small $s$ yields good performance, only restricted features are enough to explain the information processing in the neural networks.

\subsection{Concrete example of the proposed method}

Here, we give a concrete neural network for the proposed method. 

First, we train a neural network that solves a classification problem in advance. The classification problem is to recognize a $28\times28$ pixel handwritten digit image from the MNIST dataset as a numerical value \cite{LeCun1998,Deng2012}. The neural network is a fully connected network whose activation function is the ReLU function $h(x) = \max\{0, x\}$. The number of nodes in each layer is $(784, 20, 20, 20, 20, 20, 10)$, where the first is the input layer, the last is the output one, and the rest correspond to multiple intermediate layers. The activation function is not applied to the first intermediate and output layers; all nonlinear processing is performed inside the other intermediate layers. The settings for training the neural network are shown in table~\ref{tab:learn_settings}.

\begin{table}[tb]
    \begin{center}
    \caption{Training settings of the neural network.}
    \label{tab:learn_settings}
    \begin{tabular}{c r}
        \hline
        number of epochs & 14 \\
        optimization algorithm & \verb|AdaDelta| \\
        hyper parameter for \verb|AdaDelta| $\rho$ & 0.9 \\
        initial learning rate & 1.0 \\
        decay of learning rate & 0.7 times per epoch \\
        \hline
    \end{tabular}
    \end{center}
\end{table}

\begin{figure}[tb]
\centering
\includegraphics[width=90mm]{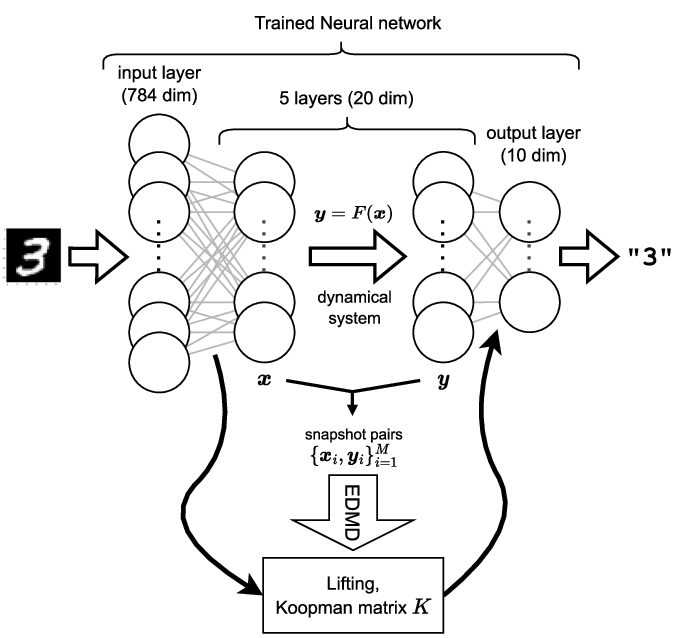}
\caption{Neural network, discrete-time dynamical system, and the partial replacement with the Koopman matrix.}
\label{fig:network}
\end{figure}

After training the neural network, we compute the Koopman matrix that mimics the internal processing of the network using EDMD. The EDMD algorithm requires snapshot pairs of the dynamic system. It is impossible to directly pair the inputs and outputs of the neural network of interest since the snapshot pairs must have the same dimension. Therefore, we consider the inputs and outputs of two layers in the intermediate layers as the snapshot pairs of the discrete-time dynamical system; they have the same dimension when training data are forwardly propagated to the trained neural network. In the experiment, we select the first and last layers of the intermediate layers to generate snapshot pairs. Then, the Koopman matrix mimics all the nonlinear processing of the neural network. Figure~\ref{fig:network} shows an overview of the network structure and application of EDMD.

\subsection{Preliminary investigation of nonlinearity with the TT format}

We investigate the roles of nonlinearity in the trained neural network, that is, to what extent the neural network partially replaced by the Koopman matrix replicates the original behavior of the intermediate layers.

We consider the following monomial dictionary:
\begin{eqnarray}
\phi: \bm x \mapsto \prod_{d=1}^D x_d^{n_d},
\end{eqnarray}
where $0 \le n_d \le N_{\mathrm{max}}$ for $d = 1, 2, \dots, D$. The intermediate layers in the neural network in figure~\ref{fig:network} has $20$ nodes. Hence, even when $N_{\mathrm{max}} = 2$, the dictionary size is $L = 3^{20} = 3,486,784,401$. We obviously cannot create the Koopman matrix with this dictionary size. Hence, the TT format is employed. We here give only a brief review of the TT format; see, for example, \cite{Oseledets2009a,Oseledets2009b,Oseledets2011} for the details of the TT format. 

The TT format for a tensor $\mathbf{T} \in \mathbb{R}^{n_1 \times \cdots \times n_D}$ is expressed as
\begin{eqnarray}
\mathbf{T} = \sum_{k_0=1}^{r_0} \cdots \sum_{k_D=1}^{r_D}
\mathbf{T}_{k_0,:,k_1}^{(1)} \otimes \cdots \otimes \mathbf{T}_{k_{D-1},:,k_{D}}^{(D)},
\end{eqnarray}
where $r_0, \dots, r_D$ are called TT ranks, and $r_0 = r_D = 1$. The tensors of order $3$, $\mathbf{T}^{(d)} \in \mathbb{R}^{r_{d-1} \times n_d \times  r_d}$, are called TT cores. The following formal expression for the TT core would be helpful for understanding. That is, imagine a matrix whose element is a vector as follows:
\begin{eqnarray}
\left[ \mathbf{T}^{(d)} \right]
= \left[ 
\begin{array}{ccc}
\mathbf{T}_{1,:,1}^{(d)} & \dots & \mathbf{T}_{1,:,r_d}^{(d)} \\
\vdots & \ddots & \vdots \\
\mathbf{T}_{r_{d-1},:,1}^{(d)} & \dots & \mathbf{T}_{r_{d-1},:,r_d}^{(d)} 
\end{array} \right],
\end{eqnarray}
where $\mathbf{T}_{i,:,j}^{(d)}$ means the vector.

It is possible to construct the TT format for the dictionary vector $\bm{\Phi}(\bm{x}_t) \in \mathbb{R}^{n_1 \times \cdots \times n_D}$; in \cite{Nuske2021}, the algorithm with higher-order CUR decomposition was given. We here employ an algorithm based on the singular value decomposition, which is implemented in \verb|scikit_tt| \cite{scikit_tt}. In our experiments, the rank for each singular value decomposition is truncated so that the cumulative contribution of the singular values is over 0.999. Note that we must calculate $\bm{\Phi}(X)^\dagger$, i.e., not for a data point $\bm{x}_t$ but for the data matrix $X$. Hence, adding a core indicating the data number, we compress $\bm{\Phi}(X) \in \mathbb{R}^{n_1 \times \cdots \times n_D \times n_{D+1}}$ in the TT format, where $n_{D+1} \in \{ 1,2,\dots,M\}$. The TT format of $\bm{\Phi}(Y)$ is calculated in the same way.

It is not enough to calculate $\bm{\Phi}(X)^\dagger$ and $\bm{\Phi}(Y)$ since our final aim is to obtain the Koopman matrix $K^\top = \bm{\Phi}(Y) \bm{\Phi}(X)^\dagger$ and to predict $\bm{y}_t = F(\bm{x}_{t})$ from $\bm{x}_t$. Although the TT format for the Koopman matrix is obtained in \cite{Gelß2019}, the prediction step needs some techniques even for the TT format to reduce the memory size; we explain them in Appendix~A.

Here, we use $M = 10,000$ snapshots to construct the Koopman matrix $K$. We perform numerical experiments by changing the maximum degree $N_{\mathrm{max}}$ in the dictionary to $1, 2, 3$ and $4$. Since the number of variables is $D=20$, the dictionary size for $N_{\mathrm{max}} = 4$ is $5^{20} \simeq 9.5 \times 10^{13}$ for the naive construction. Hence, the number of elements of Koopman matrix becomes $5^{40} \simeq 9.1 \times 10^{27}$, which memory size is about $7.2 \times 10^{19}$ GB for double-precision floating-point format. By contrast, in the TT format, the final memory size for the Koopman matrix was about $1.5$ GB.

\begin{figure}[tb]
\centering
\includegraphics[width=80mm]{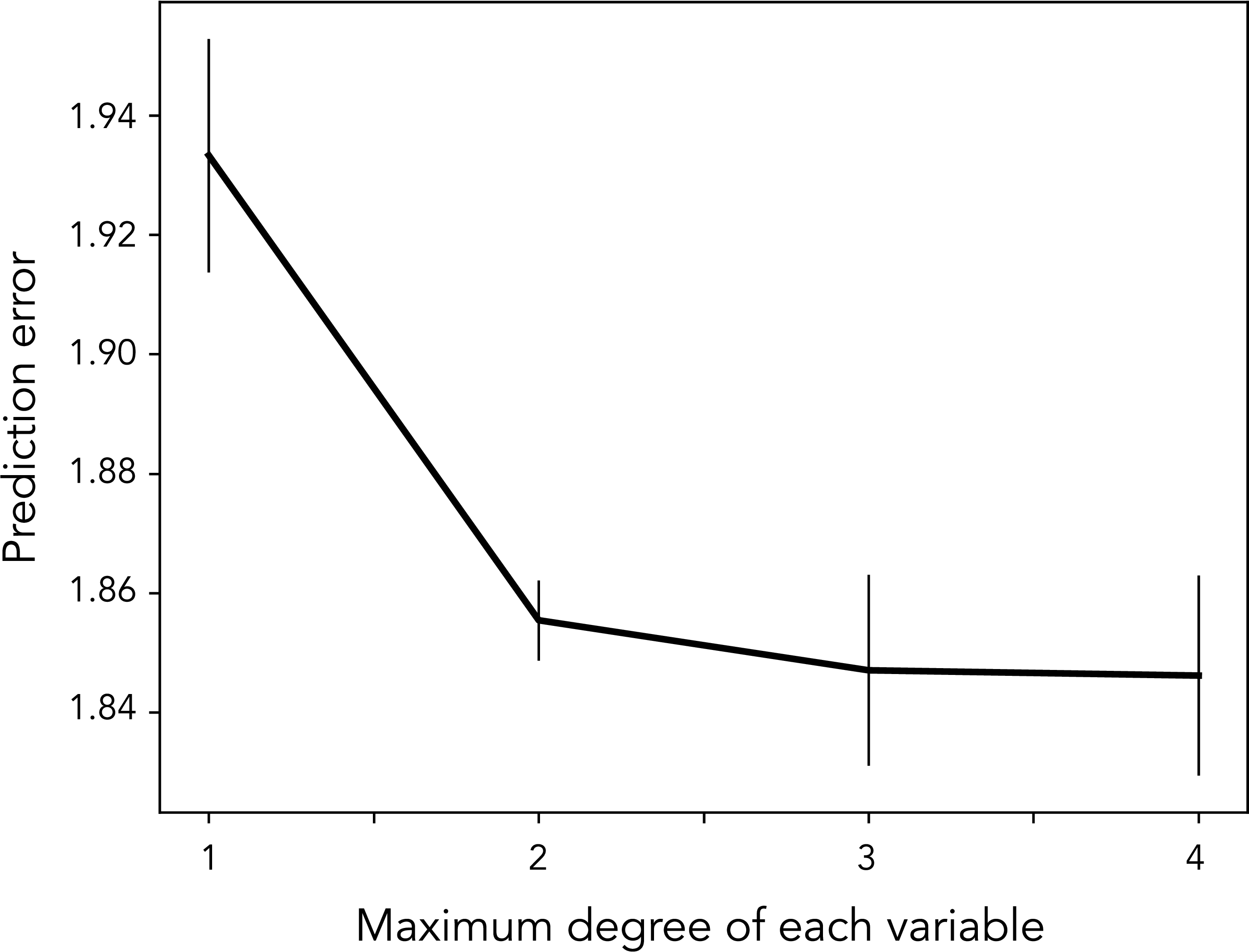}
\caption{Prediction errors for $100$ test points. Error bars correspond to the standard deviation for $5$ different experiments. The horizontal axis is the maximum degree for each variable, $N_{\mathrm{max}}$, in the dictionary.}
\label{fig:TT_errors}
\end{figure}

After constructing the Koopman matrix, we take $100$ test data and evaluate the prediction errors. Repeating the whole steps with $5$ different random seeds, the standard deviation for the prediction errors in the sense of the $L_2$ norm is evaluated. Figure~\ref{fig:TT_errors} gives the numerical results, in which the error bar means the standard errors for the $5$ repeated experiments. From figure~\ref{fig:TT_errors}, we see that the setting with $N_{\mathrm{max}} = 3$ roughly yields enough predictions. These results suggest that higher-order nonlinearity is not necessary for the neural network for this classification problem. Of course, the dictionary includes higher order functions as monomials; for example, the term $x_1^{2} x_2^{2} \dots x_{20}^{2}$ is included for the case with $N_{\mathrm{max}} = 2$. However, one would expect that restricted nonlinearity could be enough for constructing the neural network for this problem. We will discuss this point below.

From these preliminary numerical experiments with the TT format, we expect that the Koopman operator extracts the essential features of neural networks well.

\subsection{Practical dictionary for more detailed experiments}

Although the TT format is a powerful tool for high-dimensional systems, it is still difficult to apply it naively. In our naive code, a prediction per data takes about $0.7$ seconds for the TT format with $N_{\mathrm{max}} = 3$. The TT format tells us that the limited nonlinearity could be enough. Hence, we use the more limited dictionary in the following discussion.

For numerical experiments without the TT format, we consider two cases of dictionary functions for the EDMD algorithm. One is the monomial dictionary $\{\phi : \bm x \mapsto \prod_{i=1}^D x_i^{n_i} \mid 0 \le \sum_{i=1}^D n_i \le N_{\max}^{\mathrm{total}} \}$, specified with a constant $N_{\mathrm{max}}^{\mathrm{total}}$. Note that $N_{\mathrm{max}}^{\mathrm{total}}$ is different from $N_{\mathrm{max}}$ in the TT format; $N_{\mathrm{max}}^{\mathrm{total}}$ is the maximum of the \textit{total} degrees of the monomial exponents for each variable. For example, the dictionary here does not have $x_1^2 x_2^2$ for $N_{\max}^{\mathrm{total}} = 3$ since the total degree of this term is $4$.

In addition to the simple monomial dictionary, we use the Gaussian radial basis function (RBF) dictionary (including identity mapping) as the second case. The center point $\bm{c}$ of the Gaussian RBF $\phi(\bm x) =\exp(-\varepsilon \|\bm x - \bm c\|^2)$ is randomly chosen from the input of one of the snapshot pairs. The parameter $\varepsilon$ is set to $0.001$.

\begin{table}[tb]
\caption{Accuracy of networks in which the intermediate layers are replaced by the corresponding time evolution process with Koopman matrices.}
\label{tab:koopman_mimic_results}
\centering
\begin{tabular}{cr|c}
Dictionary & \multicolumn{1}{c|}{Number of dictionary functions} & Accuracy \\ \hline
Monomials up to 1st order & 21 & 79.71\% \\
Monomials up to 2nd order & 231 & 93.54\% \\
RBF & 231 & 94.69\% \\
Monomials up to 3rd order & 1771 & 95.67\% \\
RBF & 1771 & 96.01\% \\ \hline \hline
\multicolumn{2}{c|}{Original trained neural network}  & 95.62\%
\end{tabular}
\end{table}

\begin{table}[tb]
\caption{Prediction errors of the time evolution of the state variables by the Koopman matrices.}
\label{tab:koopman_mimic_results_ss}
\centering
\begin{tabular}{cr|c}
Dictionary           & \multicolumn{1}{c|}{Number of dictionary functions} & Error  \\ \hline
Monomials up to 1st order & 21 & 6.6181 \\
Monomials up to 2nd order & 231 & 3.2294 \\
RBF & 231 & 2.3843 \\
Monomials up to 3rd order & 1771 & 2.0043 \\
RBF & 1771 & 1.3154
\end{tabular}
\end{table}

After learning neural networks, we extract a data set for which the trained network successfully recognizes the image for the MNIST data. The number of data is $58,052$. Then, we construct the Koopman matrix using these data. After that preparation, we choose $10,000$ data pairs from the MNIST test data set, and snapshot pairs are generated from the intermediate layers. We finally evaluate the performance of the constructed Koopman matrix for this test data set. Table~\ref{tab:koopman_mimic_results} shows the classification accuracy for the test data with the networks substituted with the Koopman matrices for the intermediate layer. From table~\ref{tab:koopman_mimic_results}, we see that the classification problem can be solved with moderate accuracy even when the intermediate layer is replaced by time evolution using the Koopman matrix; the accuracy increases as the number of dictionary functions increases.

\begin{figure}[tb]
\begin{center}
    \includegraphics[width=110mm]{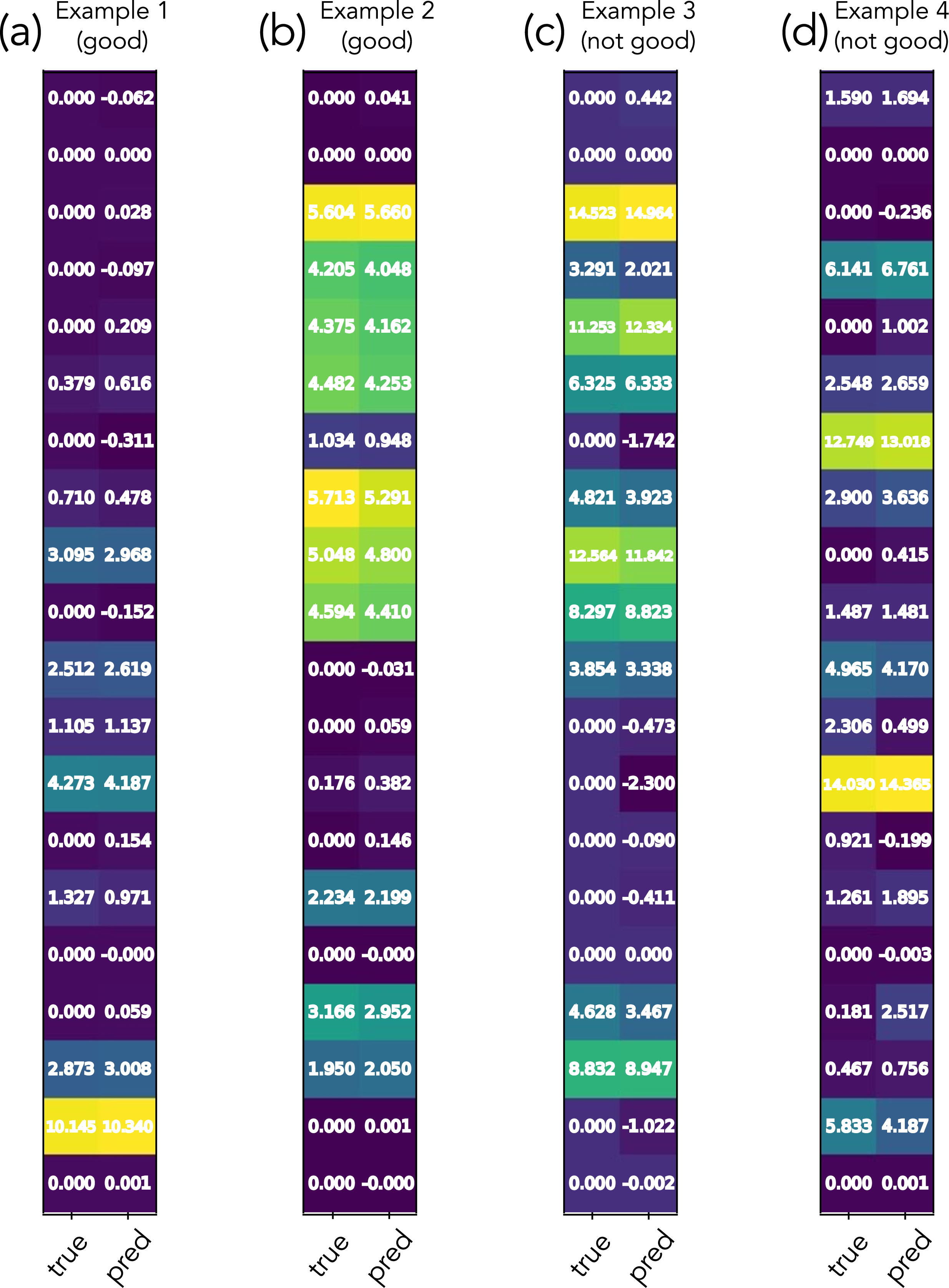}
\end{center}
  \caption{Examples of prediction for the intermediate layers. In each example, the left column shows the state variable after true time evolution, and the right column shows the prediction by the Koopman matrix. (a) and (b) corresponds to cases with successful prediction, and (c) and (d) are not-so-good ones.}
  \label{fig:ss_ga}
\end{figure}

Next, we consider the reproducibility of snapshot pairs obtained by inputting test data to the original trained neural network. Table~\ref{tab:koopman_mimic_results_ss} shows the average prediction errors; the prediction accuracy of the internal time evolution improves as the number of dictionary functions is increased, as does the accuracy of the classification. We also see that the simple dictionary construction here yields not so worse prediction errors compared with the TT format. 

Note that the prediction at the intermediate layer is not always good when we obtain a correct result in the classification task. Figure~\ref{fig:ss_ga} shows four examples of prediction of state variables. Although all four cases yield correct answers in the final output of the recognition task, the cases in figures~\ref{fig:ss_ga}(a) and \ref{fig:ss_ga}(b) show good predictions in the snapshot pair prediction stage; by contrast, several predicted elements differ from the true values in figures~\ref{fig:ss_ga}(c) and (d). Why is such a `rough' prediction enough for the classification problem? The reason is that only the maximum output in the network is taken as the final recognition stage. Hence, even if the prediction of the internal time evolution is not so accurate to some extent, the accuracy of the classification may not be affected.

\begin{figure}[tb]
\centering
\includegraphics[width=90mm]{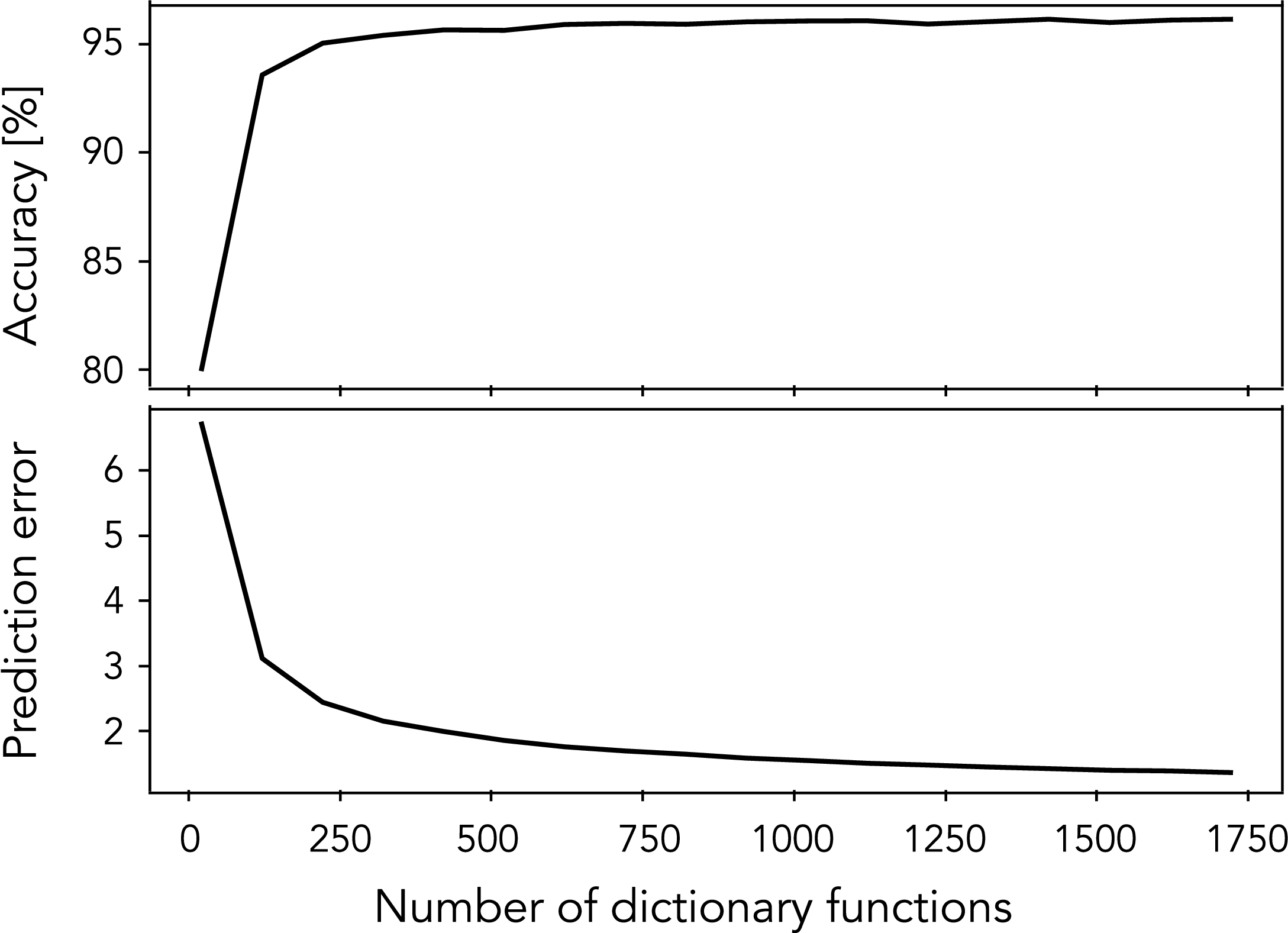}
\caption{Discrimination accuracy of the test data (top) and prediction error of the time evolution of the internal state variables (bottom) as the number of dictionary functions is increased.}
\label{fig:ss_acc_error}
\end{figure}

To confirm this in more detail, we investigate the classification accuracy of the network and the prediction error of the internal time evolution when the number of RBF dictionary functions gradually increased. Figure~\ref{fig:ss_acc_error} shows the results; the classification accuracy of the network reached a plateau, while the prediction errors of the internal time evolution did not fully decrease. Therefore, as discussed above, there are cases in which the classification accuracy can be maintained even if there is a certain amount of prediction errors in the time evolution.

\section{How much does the Koopman matrix extract the essence?}

The results in the previous section indicate that we can replace the intermediate layers with the Koopman matrix. Then, we check how much information is extracted by the Koopman matrix. In order to investigate it, we evaluate the performance by compressing the matrix with singular value decomposition. If we can achieve high compression without significant performance loss, it means that the Koopman matrix contains the essential information in a compact way. Then, we compare the compression ability of the Koopman matrix with the conventional pruning method.

\subsection{Experimental details}

In the following numerical experiments, the same neural network in section~3 is employed. We here use the Gaussian RBF (and identity mapping) dictionary, which makes it easier to adjust the number of dictionary functions than the monomial dictionary. Parameters and selection of center points for the Gaussian RBF are the same as in section~3.

It is possible to adjust the degree of the model compression of the proposed method by changing the number of ranks of the low-rank approximation and that of dictionary functions. As denoted in section~3.1, the low-rank approximation is performed by the singular value decomposition:
\begin{equation}
A = KB \simeq U \Sigma V^{\top},
\label{eq:approximate_svd}
\end{equation}
where $U \in \mathbb{R}^{L\times L}$, $\Sigma \in \mathbb{R}^{L \times D}$, and $V \in \mathbb{R}^{D \times D}$. Note that the dimension of variables $D=20$ in the whole experiments. We select only $s$ values from the singular values $\sigma_1, \ldots, \sigma_s > 0 \ ( s \leq \min\{L,D\})$. Hence, a lower number of ranks $s$ means a highly compressed matrix.

If the number of ranks $s$ of the low-rank approximation is too large, the total number of the matrix elements in the three matrices of the singular value decomposition will increase more than that of the original matrix. Therefore, we experiment on the case without the low-rank approximation and the case with a sufficiently small number of ranks. The number of dictionary functions $L$ was tested in each case from $21$ to $120$.

For comparison, we use the naive conventional pruning method. Although the neural network is the same as in section~3, it is compressed by the pruning, as follows. Pruning is a method of reducing parameters by selecting unimportant weights and removing them. There are mainly two types of pruning: unstructured pruning and structured pruning. Unstructured pruning only sets some weights to 0, which does not reduce the memory used by the network to perform inference, but it is likely to have high accuracy. By contrast, in structured pruning, weights are removed in units of rows, columns, channels, and so on. Hence, we can reduce the size of the network structure and the memory size required to run the inference while it gives less accuracy than unstructured pruning.

In this experiment, we test both unstructured pruning and structured pruning and compare with the proposed method. There are several criteria for selecting the unnecessary weights. Here, we adopt magnitude-based pruning, in which the weights with the small $L_1$ norms are removed.

\subsection{Experimental results}

\begin{figure}[tb]
\centering
\includegraphics[width=100mm]{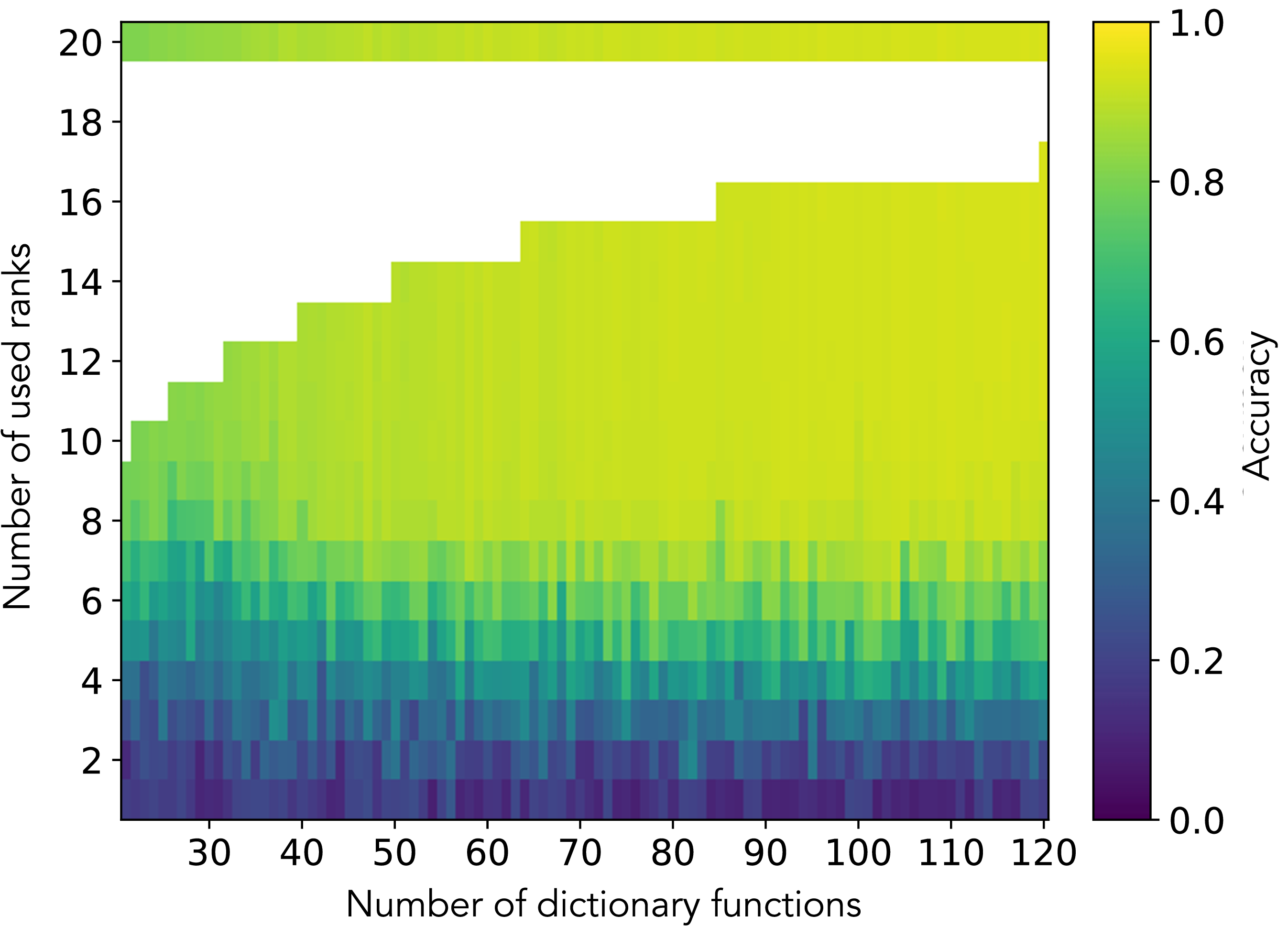}
\caption{Accuracy of the compressed model by the proposed method for various settings. In the blank region, the number of matrix elements is greater than that of the case without singular value decomposition.}
\label{fig:comp}
\end{figure}

\begin{figure}[tb]
\centering
\includegraphics[width=90mm]{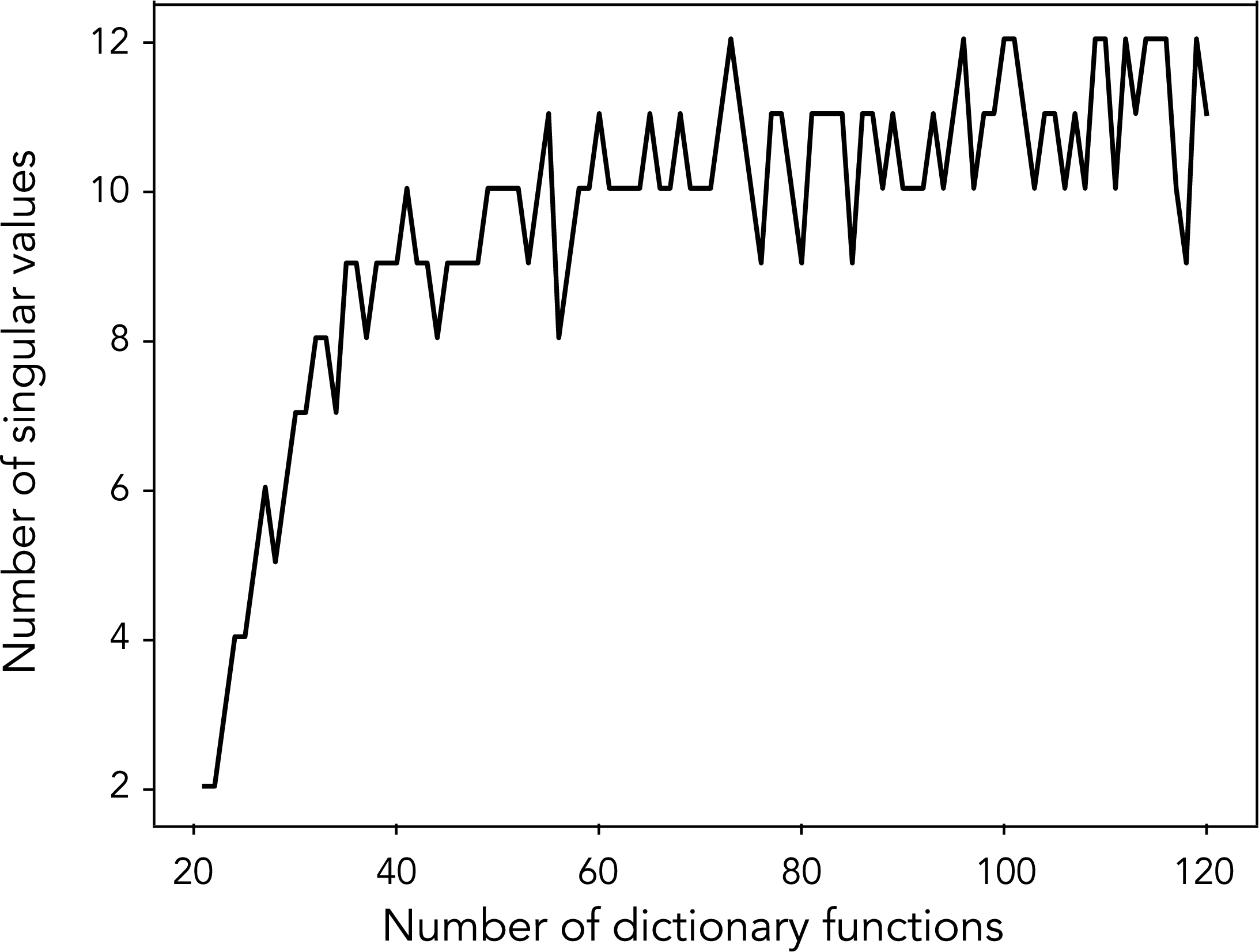}
\caption{Number of singular values (vertical axis) at the point when the cumulative contribution of the singular values exceeds 99\%. The horizontal axis is the number of dictionary functions.}
\label{fig:singular_val}
\end{figure}

Figure~\ref{fig:comp} shows the accuracy of test data of the neural network with model compression in each setting: the number of dictionary functions and the number of ranks, which are shown in the vertical and horizontal axes, respectively. Note that we plot the results without singular value decomposition for the uppermost part of figure~\ref{fig:comp}. When the number of ranks is 20, which equals the dimension of the state variable, the singular value decomposition only increases the number of matrix elements and does not change the calculation results. Hence, the results without performing singular value decomposition are shown for the 20-rank cases. The blank space corresponds to the region where the total number of matrix elements in the three matrices of the singular value decomposition are greater than that of the case without singular value decomposition.

From figure~\ref{fig:comp}, we see that the accuracy increases gradually as the number of dictionary functions increases. Note that we achieve roughly the same accuracy in the region where the ranks are greater than 10 regardless of the number of dictionary functions. We plot the number of singular values with which the cumulative contribution of singular values exceeds 99\% in figure~\ref{fig:singular_val}. These results indicate that about 10 singular values are enough to approximate the behavior of the intermediate layer.

\begin{figure}[tb]
\centering
\includegraphics[width=90mm]{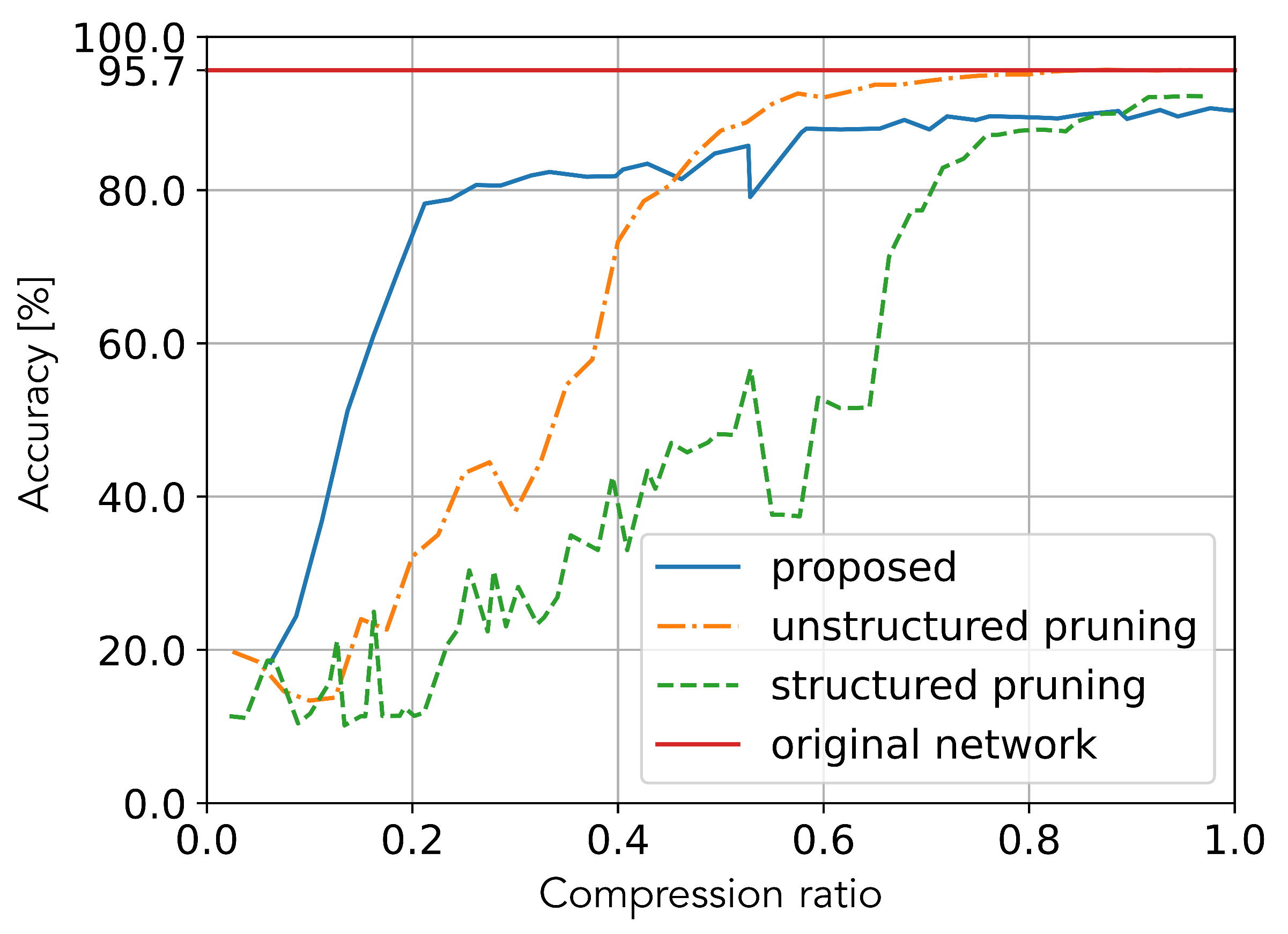}
\caption{Accuracy for various compressed models against compression ratio which is proportional to the number of parameters in the intermediate layer of the network.}
\label{fig:comp_rate}
\end{figure}

\begin{figure}[tb]
\centering
\includegraphics[width=100mm]{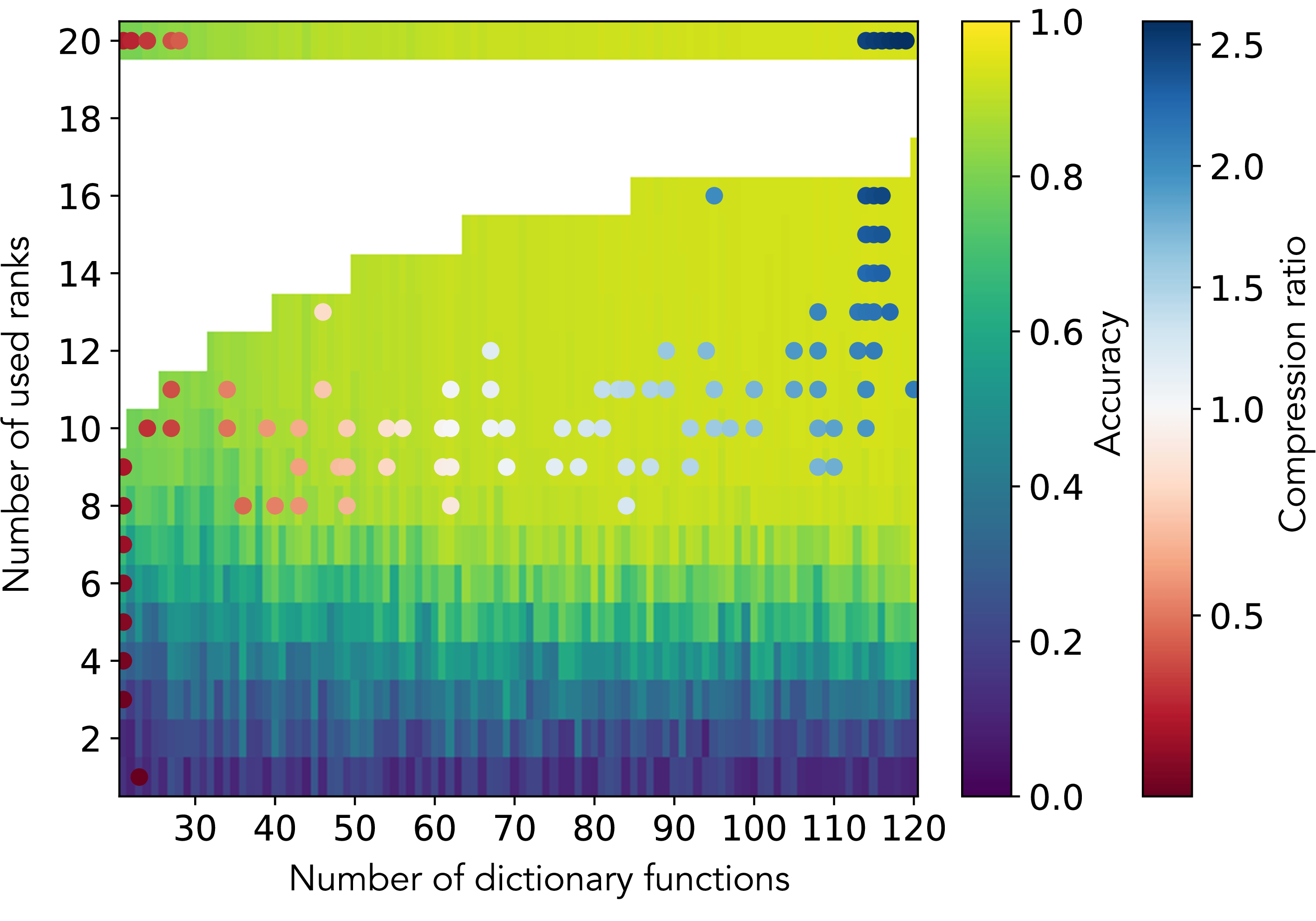}
\caption{Accuracy on the validation data and the settings adopted for each compression ratio.}
\label{fig:comp_scat}
\end{figure}

Figure~\ref{fig:comp_rate} shows the accuracy of the test data for various compression ratios. The vertical axis represents the accuracy, and the horizontal axis represents the compression ratio. In this work, we define the compression ratio as the ratio of \textit{the number of parameters in the intermediate layer of the network after compression} to \textit{those of the original network}. Hence, the smaller compression ratio means a higher compressed network. Note that the parameters of the network compressed by the proposed method include the elements of the Koopman matrix and the variables needed to represent the center points of Gaussian RBFs used as dictionary functions. The compression ratio is evaluated with the number of the whole parameters.

To plot the accuracy of the proposed method for each compression ratio in figure~\ref{fig:comp_rate}, we determine the number of dictionary functions and ranks in the following procedure. First, the training data is split: 20\% as validation data and the rest used to compress the network under various settings. Second, we seek the most accurate setting on the validation data among those with the same compression ratio. Figure~\ref{fig:comp_scat} shows the accuracy of the proposed method on the validation data with the scatter plots corresponding to the best settings for various compression ratios. The accuracy values on these points in the scatter plots were used to plot figure~\ref{fig:comp_rate}.

When the number of parameters in the network is less than half of those in the original network, the intermediate layer replaced by the Koopman matrix results in higher accuracy than the two conventional compression methods. Indeed, figure~\ref{fig:comp_rate} shows that the accuracy of the proposed method reaches about 80\% in the range of a compression ratio of $0.2$ to $0.4$. This fact means the restricted nonlinearity replaces the intermediate layer of the network well. Actually, figure~8 supports this fact. There are two types of scatter plots for cases with compression ratios roughly ranging from $0.2$ to $0.4$:
\begin{enumerate}
\item The number of dictionary functions is small, while the rank is 20; see the left-upper region in figure~\ref{fig:comp_scat}.
\item The rank is around $10$, while the number of dictionary functions is not so small; see the light green region in figure~\ref{fig:comp_scat}. 
\end{enumerate}
Hence, a small number of substantial dictionary functions is enough to mimic the intermediate layer.

In addition, we perform additional numerical experiments for the Fashion MNIST dataset\cite{Xiao2017}. The numerical results show the same behavior; see the numerical results in Appendix B. These results indicate that the Koopman matrix can grasp the essence of the intermediate layers in the neural network in a compact way; a small part of the elements contains the crucial information of the intermediate layers. 

The purpose of this paper is not to obtain the best pruning performance. However, some readers may be interested in the pruning performance. Then, we show some numerical results for fine-tuning cases in Appendix C.

\section{Conclusion}

In this study, we investigated the ability of the Koopman operator approach to investigate the neural networks. We performed the replacement of the intermediate layers on a network for recognition tasks with the Koopman matrix. The experimental results show that the Koopman matrix achieves good performance even in high compression cases. Hence, the essential parts of the intermediate layers are included in the Koopman matrix compactly.

There is room for further study in the case of other network structures and different compression methods of matrices. Of course, we expect the Koopman operator to extract useful information from neural networks, and further studies are hopeful. Our approach is based on the theory of dynamical systems, and there are many studies on this topic, as introduced in section~1. Then, it is hoped that the approach with the dynamical system will advance our understanding of neural networks from a physics perspective. We are performing some preliminary experiments in our ongoing research. The experiments indicate that we can replace some parts of the not-so-large neural networks with Koopman matrices, as shown in the current work. Then, we must tackle high-dimensional problems that stem from large neural networks in the future. We believe that the Koopman operator theory and the TT format used in this work could help future works that provide insights into the internal mechanisms of black-box neural networks.

\ack
This work was supported by JST FOREST Program (Grant No.~JPMJFR216K, Japan).

\appendix

\section{Construction and usage of Koopman matrix in the TT format}

\begin{figure}[th]
\centering
\includegraphics[width=90mm]{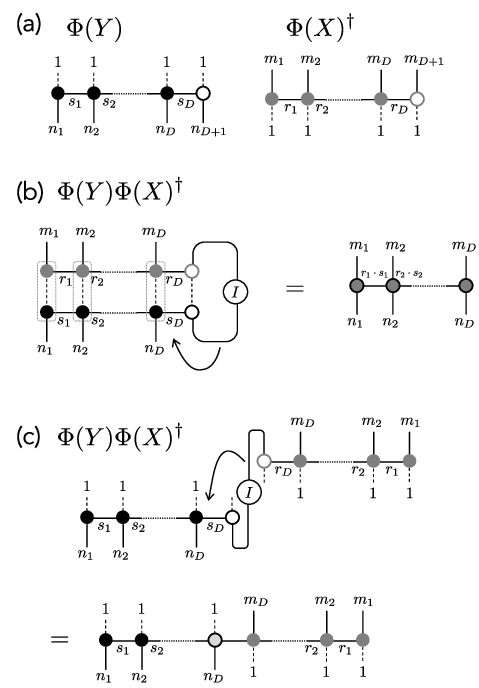}
\caption{Two ways of the construction of Koopman matrix in the TT format. (a) shows the TT formats for $\bm{\Phi}(Y)$ and $\bm{\Phi}(X)^\dagger$. (b) is the naive construction of the Koopman matrix, whose TT cores are constructed by matrices, not vectors. (c) is another construction of the Koopman matrix, in which vectors are elements of the TT cores.}
\label{fig:appendix_MPO}
\end{figure}

\begin{figure}[th]
\centering
\includegraphics[width=90mm]{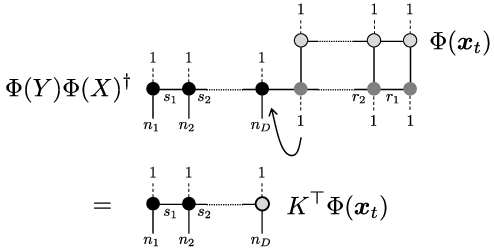}
\caption{TT format for the prediction. We finally obtain $K^{\top} \bm{\Phi}(\bm{x}_t) = \bm{\Phi}(Y) \bm{\Phi}(X)^\dagger \bm{\Phi}(\bm{x}_t)$.}
\label{fig:appendix_apply_Koopman_with_TT}
\end{figure}

In \cite{Gelß2019}, the Koopman matrix is constructed with the TT format; the constructed Koopman matrix is compared to exact solutions. For our purpose, we need to make predictions for the snapshot pairs. We need some simple techniques for this purpose; here we yield them with the diagram notations.

The Koopman matrix is a kind of operator. As denoted in section~3, the TT formats for $\bm{\Phi}(Y)$ and $\bm{\Phi}(X)^\dagger$ are constructed from data. Figure~\ref{fig:appendix_MPO}(a) shows these TT formats. Here, we also depict the dashed line with $1$, which means that a column vector in the TT core is considered as a $N_{\max} \times 1$ matrix, not a vector. Actually, the implementation in \verb|scikit_tt| \cite{scikit_tt} obeys this interpretation of the TT cores. Then, a naive construction of the Koopman matrix should be the form in figure~\ref{fig:appendix_MPO}(b). Note that the ranks are given as the multiplications of ranks of $\bm{\Phi}(Y)$ and $\bm{\Phi}(X)^\dagger$. In our numerical experiments with the number of data $M = 10,000$, the final ranks become the order of thousands, and it is not tractable even in the TT format.

Hence, we use the different contraction for the two TT formats for $\bm{\Phi}(Y)$ and $\bm{\Phi}(X)^\dagger$, as depicted in figure~\ref{fig:appendix_MPO}(c). Clearly, the elements of TT cores is not usual matrices but vectors. However, we can avoid the multiplicative increase of the ranks, and the prediction is also possible as depicted in figure~\ref{fig:appendix_apply_Koopman_with_TT}. Numerical results in section~3.3 were obtained with this techniques of the TT format for the Koopman operator.

\section{Experiments for the Fashion MNIST data set}

\begin{figure}[tb]
\centering
\includegraphics[width=100mm]{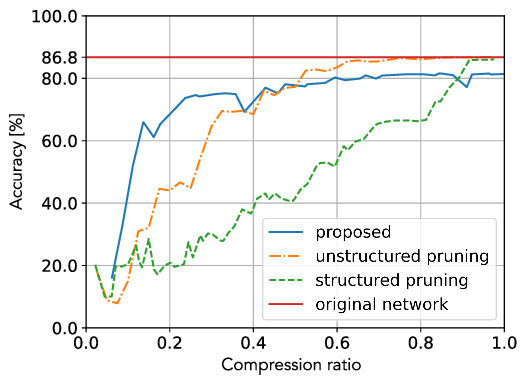}
\caption{Accuracy for various compressed models on the Fashion MNIST data set. The corresponding MNIST results are shown in figure~4.}
\label{fig:fMNIST}
\end{figure}

We perform additional numerical experiments on a different dataset. Here, we apply the same procedure in section~4 on the Fashion MNIST dataset\cite{Xiao2017}. The Fashion MNIST dataset is known to be a more difficult dataset to classify than the MNIST dataset. Since the input data size of the Fashion MNIST is the same as the MNIST, it is possible to use the same network architecture. The experimental settings are exactly the same as in section~4, and the results are shown in figure~\ref{fig:fMNIST}. This figure corresponds to figure~\ref{fig:ss_acc_error} on the MNIST dataset. We see that similar behavior on the MNIST dataset, which suggests the consequences do not depend on the dataset.

\section{Comparison of performance after fine tuning}

Additional training after the model compression of a network is called fine-tuning. Fine-tuning is useful when a certain amount of computational resources are available for training. Even if a significant number of parameters are reduced by model compression, we can recover the accuracy close to the uncompressed one. 

We here compare the performance of the proposed method with the fine-tuned conventional methods. The training procedure is the same as the first training. Only the difference is the number of epochs; the number is 5 in the fine-tuning and 14 in the original training. In the following experiments, we do not consider the center point $\bm{c}$ and the parameter $\varepsilon$  for a Gaussian RBF as training parameters.

\begin{figure}[tb]
\centering
\includegraphics[width=100mm]{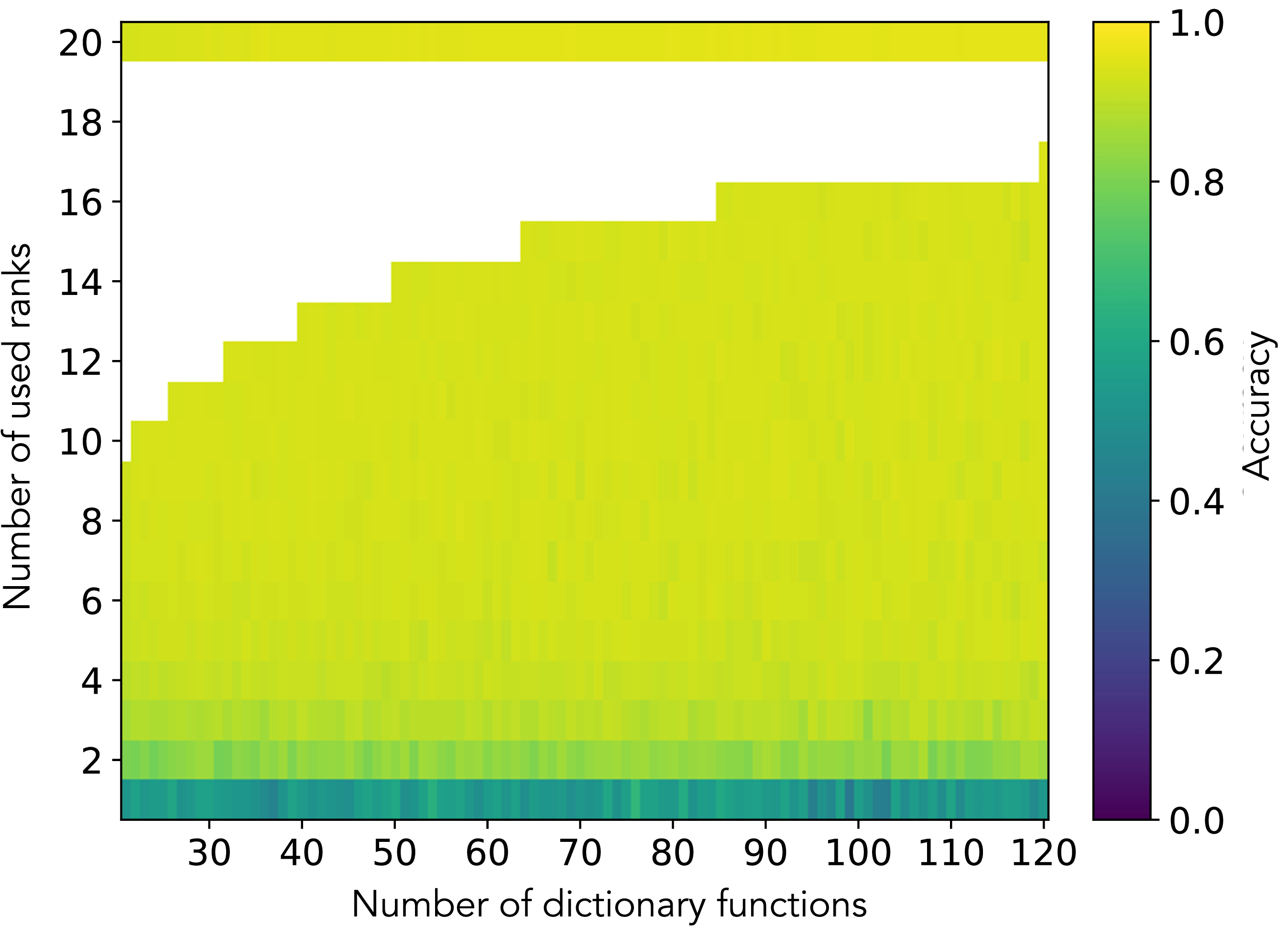}
\caption{Accuracy of fine-tuned compressed model by the proposed method.}
\label{fig:comp_finetuned}
\end{figure}

\begin{figure}[tb]
\centering
\includegraphics[width=90mm]{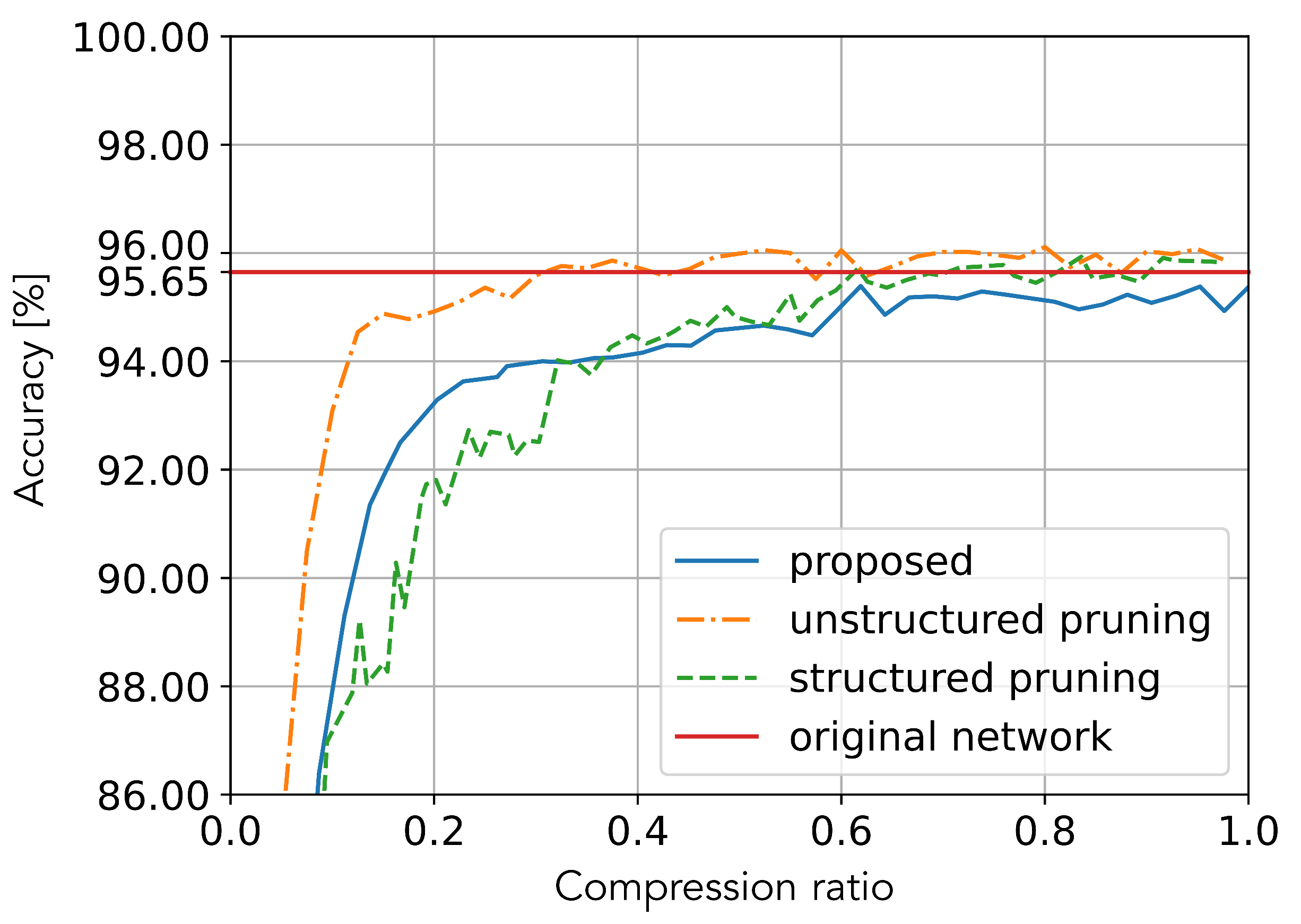}
\caption{Comparison of accuracy for various compressed models after fine-tuning.}
\label{fig:comp_rate_finetuned}
\end{figure}

\begin{figure}[tb]
\centering
\includegraphics[width=100mm]{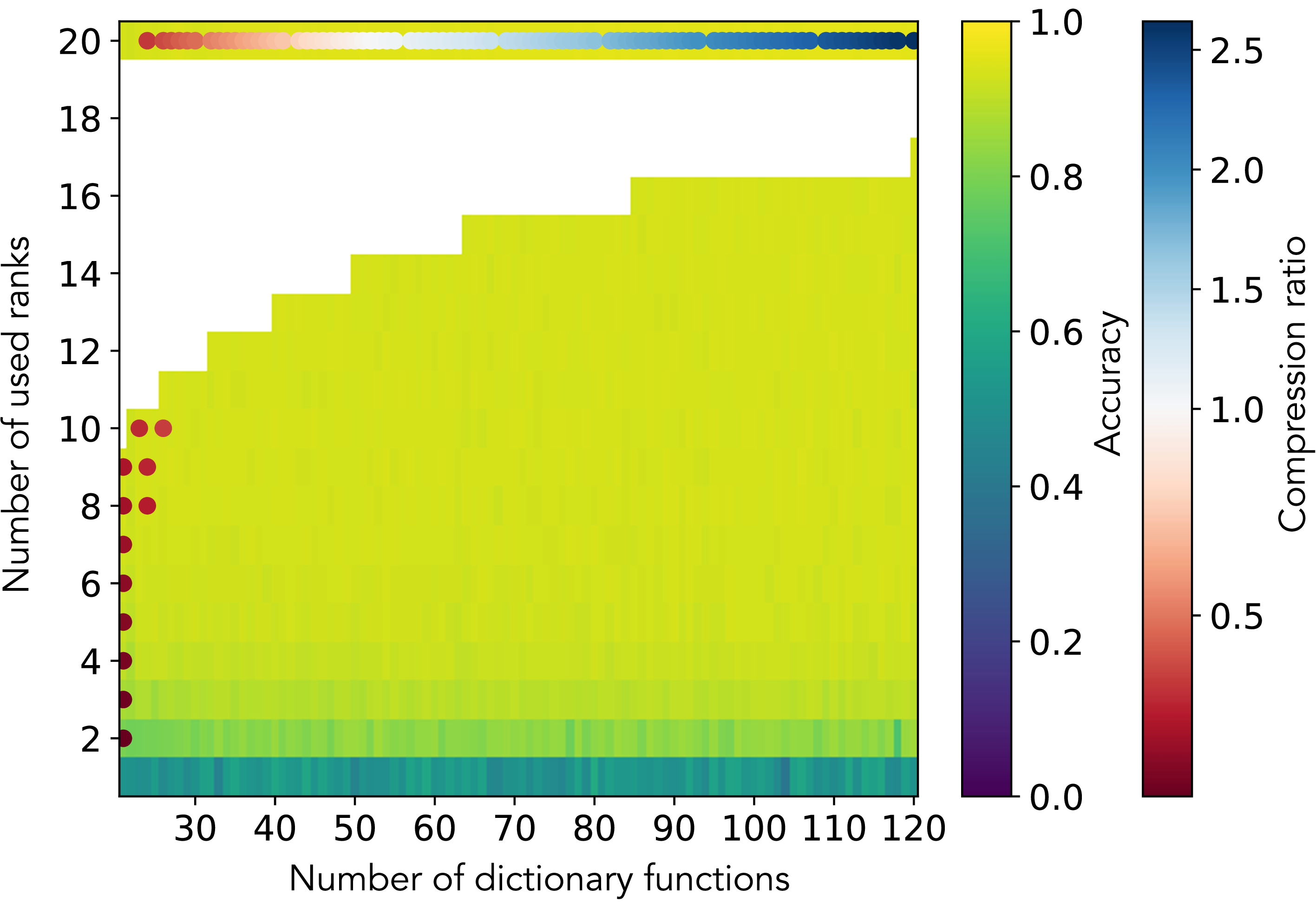}
\caption{Accuracy on the validation data after fine-tuning and the best settings adopted for each compression ratio.}
\label{fig:comp_scat_finetuned}
\end{figure}

Figure~\ref{fig:comp_finetuned} shows the accuracy of the fine-tuned compressed model by the proposed method. Figure~\ref{fig:comp_rate_finetuned} plots the accuracy versus compression ratio for the proposed and the conventional methods. Figure~\ref{fig:comp_scat_finetuned} shows the accuracy of the validation data after fine-tuning and the best settings adopted for each compression ratio, as in figure~\ref{fig:comp_scat}. 

Compared with figure~\ref{fig:comp}, we see clear improvements in accuracy in figure~\ref{fig:comp_finetuned}; the fine-tuning works well. Note that figure~\ref{fig:comp_rate_finetuned} means that the fine-tuning considerably improves the accuracy of the conventional two methods. The proposed method yields a similar accuracy with the structured pruning for region of the compression ratio smaller than $0.4$. However, the accuracy of the proposed method is lower than those of the two conventional methods, especially for the compression ratio larger than $0.4$. We understand the reason from figure~\ref{fig:comp_scat_finetuned}. That is, most of the best settings are obtained when the rank is $20$, which means no effects of the singular value decomposition. These results suggest that the structure of the Koopman matrix after singular value decomposition could not be suitable for fine-tuning.

Considering that the proposed method reduces the structure of the network itself and reduces the runtime memory as well as structured pruning, we conclude that the performance of the proposed method is comparable to that of the conventional method when the degree of compression is high.

%\bibliography{biblio}
%\bibliographystyle{unsrt}

\end{document}